\documentclass[letterpaper, 10 pt, conference]{ieeeconf}  

\makeatletter
\let\NAT@parse\undefined
\makeatother
\IEEEoverridecommandlockouts                              

\usepackage{natbib}
\usepackage{graphics} 
\usepackage{epsfig} 
\usepackage{mathptmx} 
\usepackage{times} 
\usepackage{amsmath} 
\usepackage{amssymb}  
\usepackage{times}
\usepackage{amsmath,amssymb,amsfonts}
\usepackage{xcolor}
\usepackage{hyperref}
\usepackage{verbatim}   
\usepackage{algorithm,algorithmic}
\usepackage{mathtools}
\usepackage{siunitx}
\usepackage{pgfplots}
\usepackage{pgfplotstable}
\usepackage{multirow}
\usepgfplotslibrary{fillbetween}
\usepackage{booktabs}
\usepackage{wrapfig}
\usepackage{float}
\usepackage{stfloats}
\usepackage{epstopdf}
\usepackage{siunitx}
\DeclareMathAlphabet{\mathcal}{OMS}{cmsy}{m}{n}
\DeclareSIUnit{\inch}{in}
\DeclareSIUnit{\cm}{cm}

\pgfplotsset{compat = newest}
\usetikzlibrary{arrows,positioning,shapes,intersections,external,patterns,calc,fit,decorations,decorations.markings} 
\title{\LARGE \bf
Plug-and-Play Physics-informed Learning \\using Uncertainty Quantified Port-Hamiltonian Models}

\author{Kaiyuan Tan$^{1}$, Peilun Li$^{1}$, Jun Wang$^{2}$, Thomas Beckers$^{1}$
\thanks{$^{1}$The authors are with the Department of Computer Science,
        Vanderbilt University, Nashville, TN 37212, USA.
        {\tt\small{\{kaiyuan.tan, peilun.li, thomas.beckers}@vanderbilt.edu\}}}%
\thanks{$^{2}$The author is with the Department of Electrical and Systems Engineering, Washington University in St. Louis,
Saint Louis, MO 63108, USA.
        {\tt\small \{junw@wustl.edu\}}}%
}

\newtheorem{assum}{Assumption}

\newcommand\tran{\mkern-2mu\raise1.25ex\hbox{$\scriptscriptstyle\top\hspace{0.5mm}$}\mkern-3.5mu}
\newcommand{\R}{\mathbb{R}}

\newcommand{\D}{\mathcal{D}}
\newcommand{\X}{\mathcal{X}}

\newcommand{\Z}{\mathcal{Z}}
\newcommand{\bm}[1]{{\boldsymbol{#1}}}

\newcommand{\z}{\bm z}

\newcommand{\x}{\bm x}
\newcommand{\dxdt}{\frac{\partial \x}{\partial t}}

\newcommand{\f}{\bm{f}}

\renewcommand{\u}{\bm{u}}
\newcommand{\y}{\bm{y}}

\usepackage[noabbrev]{cleveref} 
\crefname{rem}{Remark}{Remarks}
\crefname{exam}{Example}{Examples}
\crefname{assum}{Assumption}{Assumptions}
\crefname{prop}{Proposition}{Propositions}
\crefname{propy}{Property}{Properties}
\crefname{cor}{Corollary}{Corollaries}
\crefname{lem}{Lemma}{Lemmas}
\crefname{section}{Section}{Sections}
\crefname{thm}{Theorem}{Theorems}
\crefname{alg}{Algorithm}{Algorithms}
\crefname{defn}{Definition}{Definitions}
\crefname{figure}{Fig.}{Fig.}
\Crefname{figure}{Figure}{Figures}
\crefname{equation}{}{}

\begin{document}

\maketitle
\thispagestyle{empty}
\pagestyle{empty}

\begin{abstract}
The ability to predict trajectories of surrounding agents and obstacles is a crucial component in many robotic applications. Data-driven approaches are commonly adopted for state prediction in scenarios where the underlying dynamics are unknown. However, the performance, reliability, and uncertainty of data-driven predictors become compromised when encountering out-of-distribution observations relative to the training data. In this paper, we introduce a Plug-and-Play Physics-Informed Machine Learning (PnP-PIML) framework to address this challenge. Our method employs conformal prediction to identify outlier dynamics and, in that case, switches from a nominal predictor to a physics-consistent model, namely distributed Port-Hamiltonian systems (dPHS). We leverage Gaussian processes to model the energy function of the dPHS, enabling not only the learning of system dynamics but also the quantification of predictive uncertainty through its Bayesian nature. In this way, the proposed framework produces reliable physics-informed predictions even for the out-of-distribution scenarios.\\

Code + Dataset: \href{https://github.com/Beckers-Lab/PnP-PIML}{PnP-PIML LINK}

\end{abstract}

\section{Introduction}
Recent advances in robotics have enabled a wide range of applications where robots navigate through the environment, such as logistics~\citep{karabegovic2015application}, environmental surveying~\citep{vukobratovic2010robot}, and firefighting~\citep{hassanein2015autonomous}. However, robots, such as drones, are inherently vulnerable to damage from both rigid and deformable obstacles, e.g., electric wires in the air. Consequently, understanding and modeling the dynamics of these obstacles is crucial to improving the safety and reliability of robotic systems. Precise modeling of diverse obstacle dynamics enables robots to perform safe path planning, thereby minimizing collision risks and ensuring operational safety. However, the various physical properties of different objects can lead to hard-to-predict obstacle behaviors, making it difficult to establish a unified modeling framework that can accurately predict and respond to all these dynamics. For example, deformable springs and rigid sticks are hard to distinguish from appearance but follow different patterns. This disparity underscores the necessity for a universally applicable framework capable of accurately predicting a broad spectrum of obstacle dynamics.

Classical approaches to modeling unknown objects create (linear) autoregressive models that can capture the main behaviors of the system ~\citep{lydia2016linear}. They often face challenges when dealing with highly nonlinear effects, leading to solutions that may not accurately predict the trajectory of the obstacles. However, obstacles with nonlinear dynamics are quite common in real-world environments, e.g., cables on a construction site. In particular, the challenge of learning the dynamics of flexible obstacles is non-trivial due to the inherent complexity and nonlinearity of their behavior ~\citep{wei2017manipulator}. To address the limitations of traditional first principle models in capturing complex systems, data-driven methods have been proposed for state forecasting tasks, e.g. ~\citep{long2019pde,stephany2022pde}, leveraging their universal approximation properties. Especially, deep neural networks can learn complex, nonlinear patterns, enabling them to model intricate system dynamics that linear models cannot easily handle. However, purely data-driven approaches raise concerns about the efficiency, reliability, and physical correctness of the learned model~\citep{hou2013model}. The absence of physical priors in these models often leads to limited trustworthiness and other disadvantages such as vulnerability against adversarial attacks, see ~\citep{zhang2022adversarial,tan2023targeted}. As a result, there is growing interest in developing learning methods that are both trustworthy and capable of effectively modeling complex systems.

Some recent work has been proposed based on Physics-informed Neural Networks (PINNs) ~\citep{raissi2019physics, si2024initialization}. Compared to standard neural networks, PINNs are a supervised learning approach that considers physics constraints in the loss function. By explicitly adding physics conditions, PINNs ensure that the learning outcomes follow the encoded physics equations. However, there are also two drawbacks regarding this approach: First, we do not always have access to the exact underlying physics functions, making it challenging to properly train the PINNs. Secondly, even with the boost of physics priors, we are still eager to know the accuracy of the model for reliable decision-making. Hence, there is a demand for uncertainty quantification.

To address the challenge of limited knowledge of the exact underlying physical dynamics, the Port-Hamiltonian system (PHS) framework~\citep{nageshrao2015port, duong2021hamiltonian} has gained prominence due to its ability to systematically represent a generalized physics-consistent structure. Recently, PHSs have been combined with Bayesian learning for both ordinary differential equation (ODE)~\citep{beckers2022gaussian,li2024pygpphs} and partial differential equation (PDE)~\citep{tan2024physics}, enabling the learning of the Hamiltonian for physically consistent predictions without further knowledge of the underlying equations. By providing confidence intervals, regions of potential risk, particularly those arising from nonlinear obstacle movements, can be effectively modeled.

However, these generalized physics models are computationally demanding, making them slow to make predictions. Hence, we only want to use them when necessary, and, otherwise, use predictors such as neural networks that support quick inference. Thus, it is necessary to identify whether the output of an existing, inference-efficient predictor would be reliable, and, if not, to switch over to the slower but physically consistent model to learn the unknown dynamic with limited observations. Based on the data-efficient property of the Gaussian process, we propose a solution to provide physically reliable predictions in the case of out-of-distribution events, which can be directly integrated on top of any existing data-driven predictor. Composing a universally applicable prediction framework.

\textbf{Contribution:}
In this paper, we propose a novel plug-and-play physics-informed machine learning framework \textbf{PnP-PIML} to predict the motion of objects with complex dynamics. Our method aims to infuse the model with a deeper understanding of the physical laws governing the system. Unlike conventional pure-learning approaches, which may overlook conservation laws, our method utilizes Gaussian process distributed Port-Hamiltonian systems (GP-dPHS) as a framework to incorporate these physical priors as inductive bias directly into the learning process. By referring to our method as "plug-and-play", we indicate that it can be seamlessly integrated with any existing predictor. We leverage conformal prediction (CP) to detect unreliable predictions and encode data through the dPHS framework into the Bayesian learning part. In addition, the Bayesian nature of GP-dPHS also enables uncertainty quantification to justify the reliability of the given predictions. We evaluate the proposed method on the task of predicting the motion of a flexible object, namely, an oscillating string. 

\section{Preliminary}
In this section, we provide a brief overview of CP and Gaussian process distributed Port-Hamiltonian Systems.
\begin{figure*}[!b]
  \centering
    \includegraphics[width=0.21\linewidth]{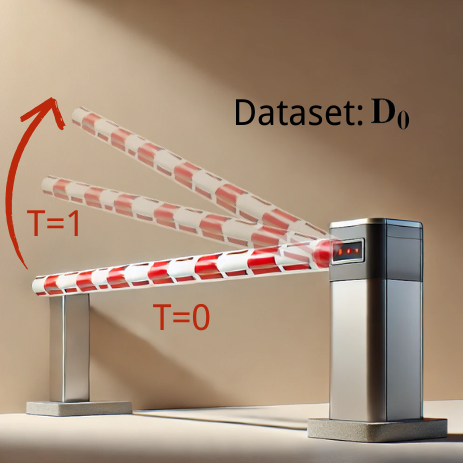}
    \includegraphics[width=0.21\linewidth]{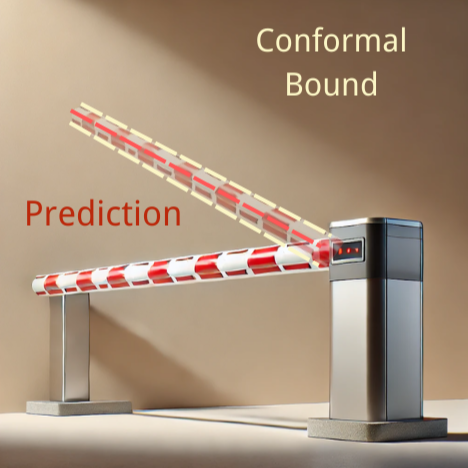}
    \includegraphics[width=0.21\linewidth]{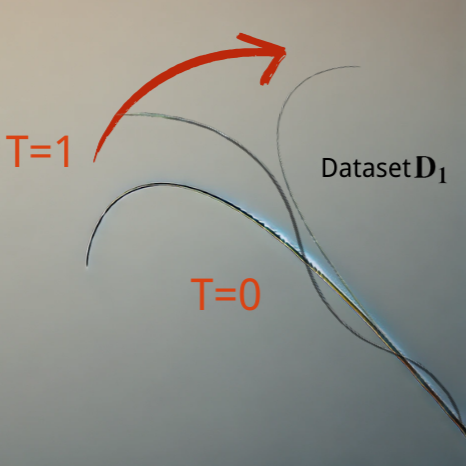}
    \includegraphics[width=0.21\linewidth]{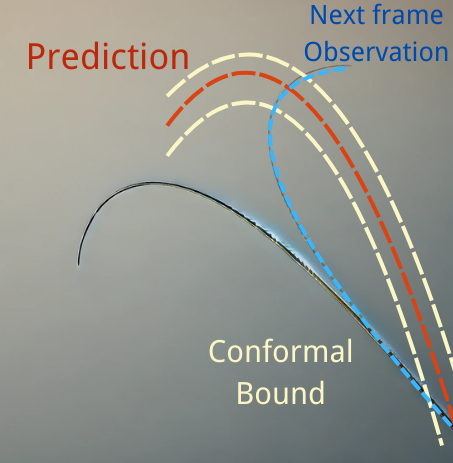}
  \caption{The left panels in the fire scene indicate that the DNN reliably predicts outcomes when test-time dynamics resemble its training regime \(\D_0\), while the right panels show that CP flags unreliable DNN outputs under OOD dynamics \(D_1\). The red arrow indicates the direction of the object movement of the future, blue denotes the ground-truth future observation, which is outbound of the CP (white).}
  \label{fig:ID_and_OOD_Visual}\vspace{-0.5cm}
\end{figure*}
\subsection{Conformal Prediction}\label{sec: CP}
Conformal prediction (CP) is a statistical technique used to quantify the uncertainty of predictions in machine learning models. It provides a prediction set that contains the true output with a user-specified probability. Unlike traditional models that give point estimates, conformal prediction gives a range or set of possible outcomes, ensuring that the correct outcome is included within that set a certain percentage of the time. We use CP to derive a non-conformity score to decide whether new test data lies within the regime of calibration data, and thus detect the existence of OOD dynamics. We briefly summarize the approach, for more details, see ~\citep{angelopoulos2021gentle,zhao2024robust}.

We formally define the out-of-distribution (OOD) detection task against in-distribution (ID) based on score \( R^{(t)} \) at time \(t\), and a constant threshold  \( C \), where the observed state data \(X\) at time step \(t+1\) can be identified as:
\begin{equation}
\begin{cases}
X_{t+1} \text{ is declared OOD if } R^{(t+1)} > C,\\[3pt]
X_{t+1} \text{ is declared ID if } R^{(t+1)} \le C.
\end{cases}
\label{eq:OOD-ID}
\end{equation}
Let \( f \) be a predictor that aims to predict trajectory \( Z^{(t+1)} \) from the input observations \( X^{(t)} \). The score \( R^{(t+1)} \) represents the prediction error, defined as \( R^{(t+1)} = \left| Z^{(t+1)} - f(X^{(t)}) \right| \), where the score \(R\) are assumed independent and identically distributed (i.i.d.) random variables following a distribution \( \mathcal{D}_0 \), where the exchangeability of \( R \) is sufficient.

In order to make a decision based on \ref{eq:OOD-ID}, we need firstly determine a constant \( C \) such that

\[
\mathbb{P}(R \leq C) \geq 1 - \delta.
\]

We can estimate \( C \) as the \((1 - \delta)\)-th quantile of the empirical distribution of \( R^{(1)}, \ldots, R^{(K)} \) among \(K\) observations in the calibration set, adjusted by a factor of \( 1/K \). Hence, with a probability more than \((1 - \delta)\), the new test data score would be less than the constant \( C \).
Formally, we set
\begin{align}\label{CP:Constant}
    C := \text{Quantile}_{(1+1/K)(1-\delta)}(R^{(1)}, \ldots, R^{(K)}),
\end{align}

which implies a lower bound on the number of calibration data points \( K \). We sort \( R^{(1)}, \ldots, R^{(K)} \) in ascending order, then $C = R^{(p)}$ corresponds to the \( p \)-th smallest nonconformity score in the calibration set. In other words, as long as the new test score \(R^{(t+1)}\) that we acquired based on the new test data is greater than the constant \( C \), we can determine it as OOD with probability \((1 - \delta)\). We visualize this OOD detection step in \ref{fig:ID_and_OOD_Visual}.

\subsection{Gaussian process Distributed Port-Hamiltonian System}\label{sec:dphs}
The composition of Hamiltonian systems through input/output ports naturally leads to the development of Port-Hamiltonian systems (PHS), a class of dynamical systems where the ports define the interactions among various components. This framework is not only applicable in the classical finite-dimensional setting but can also be extended to distributed parameter systems and multivariable cases. In the infinite-dimensional formulation, the interconnection, damping, and input/output matrices are replaced by matrix differential operators that do not explicitly depend on the state (or energy) variables. Once the Hamiltonian function is defined, the system model can be systematically derived. In fact, this generalized formulation of infinite-dimensional systems in Port-Hamiltonian form is sufficiently versatile to model classical partial differential equations and capture a broad range of physical phenomena including heat conduction, piezoelectricity, and elasticity. In the following sections, we recall the definition of dPHS as provided in~\citep{macchelli2004port}.

More formally, let $\mathcal{Z}$ be a compact subset of $\mathbb{R}^n$ representing the spatial domain, and consider a skew-adjoint constant differential operator $J$ along with a constant differential operator $G_d$. Define the Hamiltonian functional $\mathcal{H}\colon \mathcal{X} \to \mathbb{R}$ in this following form:
\[
    \mathcal{H}(\x)=\int_\mathcal{Z} H(z,x)dV,
\]
where $H\colon\Z\times\X\to\R$ is the energy density. Denote by $\mathcal{W}$ the space of vector-valued smooth functions on $\partial\mathcal{Z}$ representing the boundary terms $\mathcal{W}\coloneqq \{w\vert w=B_\mathcal{Z}(\delta_\x \mathcal{H},\u)\}$ defined by the boundary operator $B_\mathcal{Z}$. Then, the general formulation of a multivariable dPHS $\Sigma$ is fully described by
\begin{align}\label{for:pch}
\Sigma(J,R,\mathcal{H},G)=\begin{cases}
    \dxdt=(J-R)\delta_\x \mathcal{H}+G_d\u\\
\y=G_d^* \delta_\x \mathcal{H}\\
w=B_\mathcal{Z}(\delta_\x \mathcal{H},\u),
\end{cases}
\end{align}
where $R$ is a constant differential operator taking into account energy dissipation. Furthermore, $\x(t,\bm{z})\in\R^n$ denotes the state (also called energy variable) at time $t\in\R_{\geq 0}$ and location $\bm{z}\in\mathcal{Z}$ and $\u,\y\in\R^m$ the I/O ports, see ~\citep{tan2024physics} for more details. Generally, the $J$ matrix defines the interconnection of the elements in the dPHS, whereas the Hamiltonian $H$ characterizes their dynamical behavior. The constitution of the $J$ matrix predominantly involves partial differential operators. The port variables~$\u$ and $\y$ are conjugate variables in the sense that their duality product defines the energy flows exchanged with the environment of the system, for instance, currents and voltages in electrical circuits or forces and velocities in mechanical systems, see~\citep{van2000l2} for more information.

When the underlying dynamics of a system are not fully known, one can leverage the probabilistic framework provided by Gaussian processes (GPs) to model the uncertainty in the Hamiltonian function. A Gaussian process is completely characterized by its mean function and covariance function. This non-parametric Bayesian approach is particularly powerful for capturing the smooth Hamiltonian functional, especially because GPs are invariant under linear transformations~\citep{jidling2017linearly}.

Integrating these concepts, the unknown Hamiltonian function of a distributed system is encoded within a dPHS model to ensure physical consistency. Here, the unknown dynamics are captured by approximating the Hamiltonian functional with a GP, while treating the matrices $J$, $R$, and $G$ (more precisely, their estimates $\hat{J}_\Theta$, $\hat{R}_\Theta$, and $\hat{G}_\Theta$) as hyperparameters. This leads to the following GP representation for the system dynamics:
\[
    \frac{\partial \x}{\partial t} \sim \mathcal{GP}(\hat{G}_\Theta \u, k_{dphs}(\x, \x')), \label{gp:frac}
\]
with a physics-informed kernel function defined as
\[
    k_{dphs}(\x, \x') = \sigma^2_f(\hat{JR}_\Theta) \delta_\x \exp\left(-\frac{\|\x - \x'\|^2}{2 \varphi_l^2}\right)\delta_{\x'}^\top (\hat{JR}_\Theta)^\top,
\]
where $\hat{JR}_\Theta = \hat{J}_\Theta - \hat{R}_\Theta$ and the kernel is based on the squared exponential function. The training of this GP-dPHS model involves optimizing the hyperparameters $\Theta$, $\varphi_l$, and $\sigma_f$ by minimizing the negative log marginal likelihood.

Exploiting the linear invariance property of GPs, the Hamiltonian $\hat{\mathcal{H}}$ now follows a GP prior. This integration effectively combines the structured, physically consistent representation of distributed Port-Hamiltonian systems with the flexibility of GP to handle uncertainties and learn unknown dynamics from data. The resulting framework not only ensures that the model adheres to the underlying physics but also provides a comprehensive, data-informed prediction of the system's behavior.

\begin{figure*}[!b]
  \centering
    \includegraphics[width=0.7\linewidth]{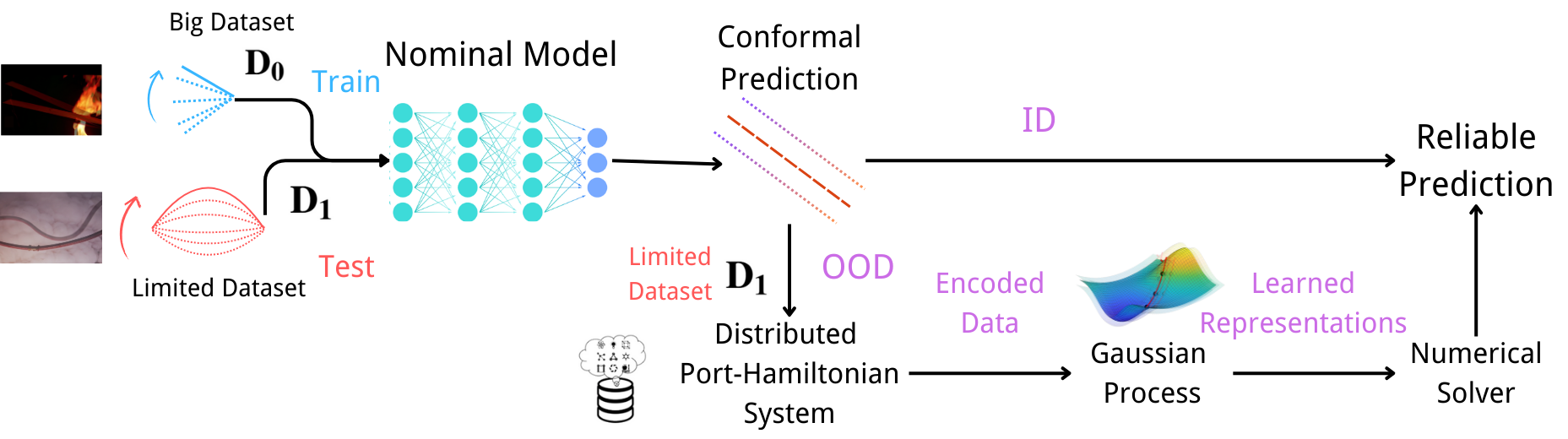}\vspace{-0.1cm}
  \caption{Illustration of proposed PnP-PIML pipeline. We leverage conformal prediction to detect outlier dynamics and, in this case, apply a data-efficient Bayesian physics-informed learning method to provide reliable predictions.}
  \label{fig:pipeline}\vspace{-0.5cm}
\end{figure*}

\section{Proposed PnP-PIML}
In this section, we will discuss the assumptions and problem formulation, followed by a detailed discussion on how the proposed physics-informed machine learning framework seamlessly integrates with existing data-driven models to enhance predictive performance.

\subsection{Assumptions and Settings}

In our case, the evolution of the states of a PDE system \( \x(t,z) \in \mathbb{R}^n \) over time \( t \in \mathbb{R}_{\geq 0} \) and spatial domain \( z \in \mathcal{Z} \) is defined by the equation with initial state \( \x(0,z) \) as described in~\ref{sec:dphs}. Give the observation of a flexible object at spatial points \(z_1\) to \(z_n\) at times \(t_1\), \(t_N\) (data set \(D_1\)), we aim to predict the future motion of the object, i.e, $\x$(\(t_{n+1}\),\(s\)). We assume that a nominal model is trained based on a data set \(D_0\) and that \(D_1\) is a different distribution as \(D_0\). When facing the unknown OOD dynamics denotes as \(D_1\), we leverage our PnP framework with the Hamiltonian structure to perform the physics-informed learning.

The Hamiltonian functional \( \mathcal{H} \in \mathcal{C}^{\infty} \) is assumed to be \textit{completely unknown} due to the unknown dynamics of the observed obstacle. Therefore, we aim to learn a dPHS model
\[
    \dxdt=(J-R)\delta_\x \hat{\mathcal{H}}+G_d\u
\]
with an estimated Hamiltonian functional \( \hat{\mathcal{H}} \) based on observations of the system~\cref{for:pch}. To effectively address the problem under consideration, the following assumptions are introduced:

\begin{assum}\label{assum:1}
We can draw \( K \) exchangeable samples from the calibration dataset of the normal learning branch, which follows the distribution \(\mathcal{D}_0\). The test time observations might be drawn from a different distribution \(\mathcal{D}_1\), where \(\mathcal{D}_0 \neq \mathcal{D}_1\).
\end{assum}

\begin{assum}\label{assum:3}
We can observe the state of the PDE system \cref{for:pch} at certain temporal points \( t_i \) and spatial points \( z_j \), resulting in a set of observations \( \{\x(t_1, z_1), \ldots, \x(t_1, z_{N_z}), \ldots, \x(t_{N_t}, z_1), \ldots, \x(t_{N_t}, z_{N_z}) \} \).
\end{assum}

\Cref{assum:1} establishes a prerequisite for the PnP setting, ensuring the presence of outlier dynamics. Additionally, it supports the validity of conformal prediction. The structure of the system matrices \( J \), \( R \), and \( G \) is assumed to be known, except for a finite set of parameters \( \Theta \subset \mathbb{R}^{n_\Theta} \), where \( n_\Theta \) is the total number of unknown parameters in these matrices. Although we assume knowledge of the dissipation matrix \( R \), we allow for unknown parameters, requiring only that the general structure be known, such as the friction model, but not the specific parameters. \Cref{assum:3} ensures that data can be collected from the PDE system, implying the state is observable. If this is not the case, an observer must be implemented.

\subsection{PnP-PIML Framework}
As shown in Fig.~\ref{fig:pipeline}, a nominal model (such as an LSTM model or a NN) is initially trained on a large-scale dataset, denoted by $\mathcal{D}_0$. Its predictive reliability is subsequently assessed on a limited observation dataset, $\mathcal{D}_1$, using conformal prediction techniques. In particular, if the computed nonconformity score is below a calibrated threshold, which means the observations are ID, the model continues to forecast future states based on the parameters learned from $\mathcal{D}_0$. Conversely, if the nonconformity score exceeds the threshold, the procedure transitions to the Bayesian physics-informed branch. This branch is designed to extract the underlying physical representations from the current limited observations $\mathcal{D}_1$, and these representations are then integrated via a numerical solver to yield accurate predictions. In the following section, the details of each step will be discussed.
\subsubsection{Data Preparation}
As described in the previous section, we define the observations as follows
\[
    \mathcal{D}_1 = \{t_i, \z_j, s(t_i, \z_j)\}_{i=1,j=1}^{i=N_t,j=N_z},
\]
where \( s(t_i, \z_j) \) denotes the observed position at time \( t_i \) and spatial point \( \z_j \), corresponding to \( N_t \) time steps and \( N_z \) spatial points. As outlined in Assumption \ref{assum:1}, a nominal predictor, represented by the function \( \f \), is trained on the dataset \( \mathcal{D}_0 \), and calibrates its conformal prediction under the same data distribution. The prediction \( P \) of the next \( w \) future frames can be formulated as:
\[
    P{(t_i, \z_j)_{i=N_{t+1},j=1}^{i=N_{t+w},j=N_z}} = \f\left(\{t_i, \z_j, s(t_i, \z_j)\}_{i=N_{t-h},j=1}^{i=N_t,j=N_z}\right),
\]
based on the \( h \) historical frames of the observations. Based on the calibration data, the conformal prediction quantile \( C \) is computed using the formulation in (\ref{CP:Constant}). If the prediction error \( R^{(i)} \) at future point \( i \) exceeds the corresponding quantile \( C \), where \( R^{(i)} \) is derived from the loss between the prediction \( P \) and real observations. It can be inferred that an outlier dynamic exists. Then, we plug in our PnP-PIML framework to perform the outlier learning (Play). The pipeline is shown in Fig.\ref{fig:pipeline}. This PnP physics-informed learning framework is applicable in scenarios where failure in the standard data-driven approach is detected. Given that data for failure cases is often scarce and time-sensitive, our Bayesian physics-informed learning method offers greater data efficiency, accuracy, and reliability compared to retraining a DNN, particularly in cases with unknown underlying physics and limited observations.

\begin{algorithm}[!t]
   \caption{PnP-PIML: Plug-and-Play Physics-Informed Machine Learning with Uncertainty Quantification}
   \label{alg:attack}
\begin{algorithmic}
   \STATE {\bfseries Input:} \{DNN training data $\bf{D}_0$, DNN $f$, test data ${\bf{D}_1}$, Distributed Port-Hamiltonian system branch $\bf{dPHS}$, Gaussian process $\bf{GP}$\}
   \STATE {\bfseries Output:} \{Reliable Prediction $\tilde{\bf{P}}$\}
   \STATE Calibrate the non-conformality score of $f$ based on data in the same distribution of  $\bf{D}_0$, get the threshold $C$
   \STATE Calculate the score $R^{new}$ for $f(\bf{D}_1)$ based on the observations
   \IF{$(R^{new} \leq C)$}
   \STATE Make prediction use existing predictor $\tilde{\bf{P}} = f(\bf{D}_1)$
   \ELSE
   \STATE Wake up $\bf{dPHS}$ branch and encode the limited observation data marked as  $\bf{D}_1'$
   \STATE Energy representation learning $E = GP(\bf{D}_1')$
   \STATE Propagate the learned representations through a numerical solver and get the prediction: $\tilde{\bf{P}} = Solver(E)$
   \ENDIF
   \STATE Output: $\tilde{\bf{P}}$
\end{algorithmic}
\end{algorithm}

We denote the temporal and spatial state variables \((p,q)\) as \(\x\), where \( p = \frac{\partial \x}{\partial t} \) and \( q = \frac{\partial \x}{\partial z} \). Recalling the dPHS formulation in \ref{for:pch}, we require the time and spatial derivatives of the state variable \(\x\). To achieve this, we apply a smooth GP function and derive the desired derivative over this function. Hence, we can acquire the desired derivative \(\frac{\partial x}{\partial t}\) for the corresponding states \(x\).

Consequently, we construct a new dataset \(\mathcal{E}\), where \(\tilde{\x}(t_i) = [\x(t_i, \z_1)^\top, \ldots, \x(t_i, \z_{N_e})^\top]^\top\) represents the spatially stacked state at time \( t_i \). And such dataset comprises the states \( X = [\tilde{\x}(t_1), \ldots, \tilde{\x}(t_{N_t})] \) and the state derivatives \( \dot{X} = \left[\frac{\partial \tilde{\x}(t_1)}{\partial t}, \ldots, \frac{\partial \tilde{\x}(t_{N_t})}{\partial t} \right] \). Thus, the dataset is defined as \(\mathcal{E} = [\dot{X}, X]\).
\begin{figure*}[!b]
  \centering
    \includegraphics[width=0.29\linewidth]{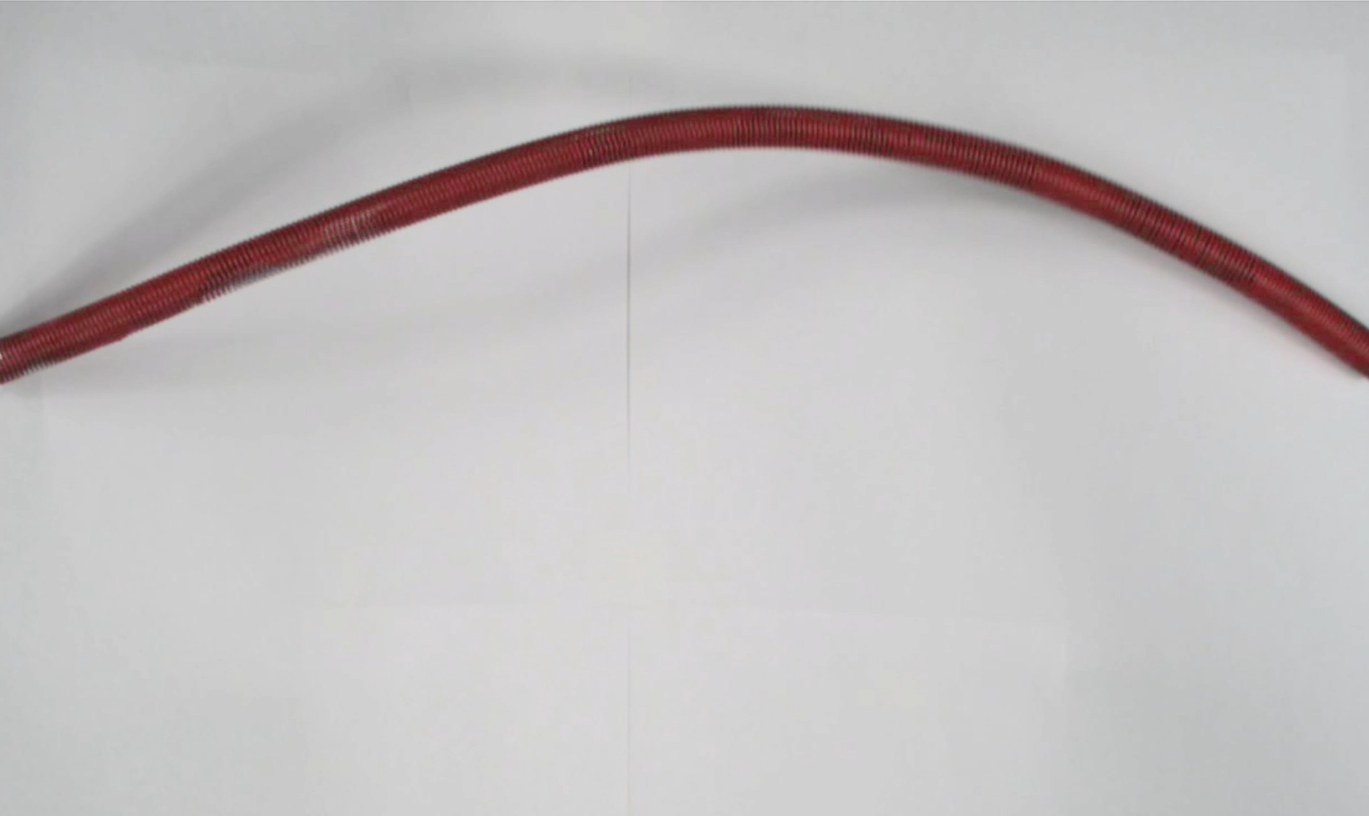}
    \includegraphics[width=0.29\linewidth]{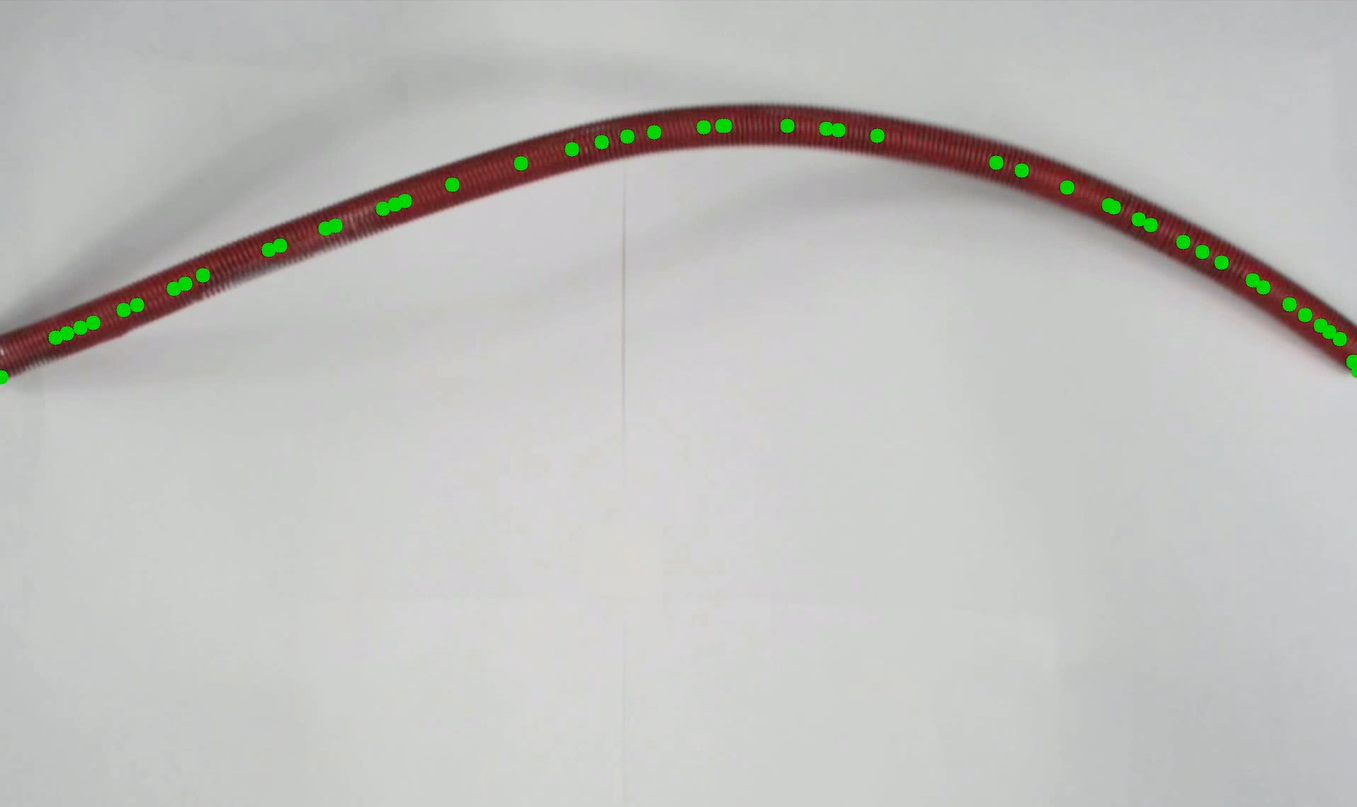}
    \includegraphics[width=0.29\linewidth]{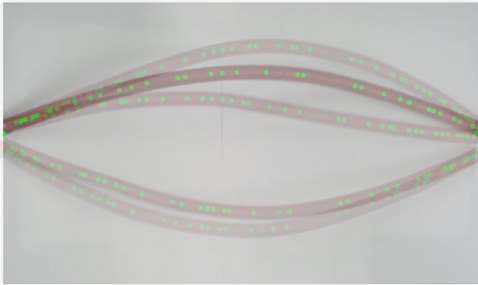}
  \caption{From left to right, the figure illustrates the process starting with the original video, skeletonization, and final denoised nonlinear movement data over time (overlay).}
  \label{fig:skeleton}\vspace{-0.5cm}
\end{figure*}
\subsubsection{Handle OOD Case}
We apply conformal prediction under calibration data distribution $\mathcal{D}_0$. Recalling the calculation of nonconformity score $R^{(i)}$ in \ref{sec: CP}, we are able to get the constant $C$ corresponds to the \( p \)-th smallest nonconformity score in the calibration set. When the score function is greater than this constant, we detect the OOD case under \((1 - \delta)\) confidence.  Whenever the OOD in dataset $\mathcal{D}_1 = \{s(t_1, z_1), \ldots, s(t_1, z_{N_z}), \ldots, s(t_{N_t}, z_1), \ldots, s(t_{N_t}, z_{N_z}) \} $ is detected, the Bayesian physics-informed learning branch is awakened (plug) and starts the learning steps (play). For many PDE physics systems, their dynamics can be encoded in distributed Port-Hamiltonian form, see \ref{for:pch}. By encoding the inputs using dPHS form, we ensure that the evolution of the dynamics follows the rules that govern the real world.

\subsection{Postprocessing after Representations Learning} \label{Data:Pre}
\subsubsection{Prediction}\label{sec: Pred}
The PnP-PIML framework aims to learn the energy representations of the system rather than directly modeling the position. 

To reduce computational complexity, we sample a deterministic Hamiltonian \(\mathcal{\hat{H}}\) from the GP distribution and use a numerical solver to compute a solution for it. Since a sample from the GP represents a deterministic function, and the dPHS structure adheres to the form in (\ref{for:pch}). This procedure ensures that the solution of the learned dynamics is physically consistent with the energy evolution, as it must satisfy the dPHS formulation.

The predicted sample is then propagated through a PDE solver with given initial conditions and time spans. Given that PDE learning is based on Bayesian physics priors, this multifaceted approach captures a powerful model of the dynamical interplay across various applications. Since its plug-and-play properties, it can be widely applied to existing learning method to deal with the outlier dynamics.

\section{Experimental evaluation}
\subsection{Setup}
 As an abstraction of deformable obstacles that robots may encounter in their operational environments, we aim to predict the motion of a Home Depot spring model \# 26455 (length of \SI{41.91}{\cm} and diameter of \SI{1.42}{\cm}) with fixed endpoints, exhibiting transversal oscillations. This setup mimics the dynamic behavior of flexible obstacles, such as a swinging electrical wire. The goal is to predict the spring’s trajectory based on test data. Initially, a pre-trained DNN is employed for prediction; however, when outlier dynamics are detected using conformal prediction, the plug-in branch is activated. We then compare the conventional data-driven retraining approach with our Bayesian physics representation learning framework under failure conditions and limited observational scenarios, thereby demonstrating the advantages of our method. The retraining process on the Lambda machine with two Nvidia GeForce RTX 4090. Moreover, to highlight the significance of incorporating the dPHS within physics-informed learning, we conduct an ablation study by comparing a baseline Gaussian Process model with our proposed approach.

\subsection{Data collection}
We utilized a high-speed camera, Blackfly S USB3 Camera Flir BFS-U3-16S2C-CS to record the motion of the spring at a frame rate of 226 frames per second (FPS). To facilitate the segmentation of the spring from the background, the spring was colored red. The initial image segmentation was performed using RGB thresholding of square errors. For this segmentation task, the RGB color range was set with a lower bound of \([150, 0, 0]\) and an upper bound of \([255, 80, 80]\). The body movement of the flexible obstacle is segmented based on the RGB range. We applied this mask to segment the spring's body. In pursuit of computational efficiency, the shape of soft obstacles is simplified using skeletonization. By applying the RGB mask, the pixel value can be described as a binary value. This step is visualized in Fig. \ref{fig:skeleton}. Considering potential video noise in real-world scenarios, we account for this by using a Kalman filter to denoise the skeletonized data. The denoised dataset, \( \hat{\mathcal{D}}_1 \), is then used to train a Gaussian Process model, with the  the spatial and temporal variables \( (t_i, z_j) \), and the state (transversal deflection) \( s(t_i, z_j) \) of the spring. By employing a smooth kernel function \( k \), such as the squared exponential kernel, the GP model is constructed. The derivatives of the GP function provide estimates for \( \frac{\partial s}{\partial t} \) and \( \frac{\partial s}{\partial z} \), allowing the dataset to be augmented with \( N_e \) additional spatial points, where \( N_e \gg N_z \). 

\subsection{Predictor structures}
\paragraph{Neural Network (nominal predictor)}
Recalling the pipeline depicted in Figure \ref{fig:pipeline}, we assume the existence of a pre-trained DNN model serving as the nominal predictor. In this study, we construct a 10-in-5-out LSTM trained on a dataset representing the oscillatory motion of a rigid body governed by near-linear dynamics. To enhance the realism of the simulated data, Gaussian white noise is introduced to the angular velocity. The model is trained on 100,000 observation samples, each consisting of 100 discrete points along the rigid body. Following training, the neural network demonstrates stable performance in predicting the dynamics of the rigid body system.

\paragraph{Gp-dPHS}
 We recast the oscillating spring system into a general dPHS framework, where, however, the Hamiltonian \(\mathcal{H}\) is unknown, leading to
 \[
     {\partial\over{\partial t}}
     \begin{bmatrix}
        p(t,z) \\
        q(t,z) \\
    \end{bmatrix}
     =
    \underbrace{ \begin{bmatrix}
        - c & {\partial\over{\partial z}} \\
        {\partial\over{\partial z}} & 0 \\
    \end{bmatrix}}_{\mathcal{J}-\mathcal{R}}
    \delta_{\x} \mathcal{H},
\]
where \(p = \frac{\partial x}{\partial t}\) and \(q = \frac{\partial x}{\partial z}\) , see~\cite{tan2024physics}.

\subsection{Results}
\begin{figure}[!b]
  \centering
  \vspace{-0.2cm}
     \includegraphics[width=0.8\linewidth]{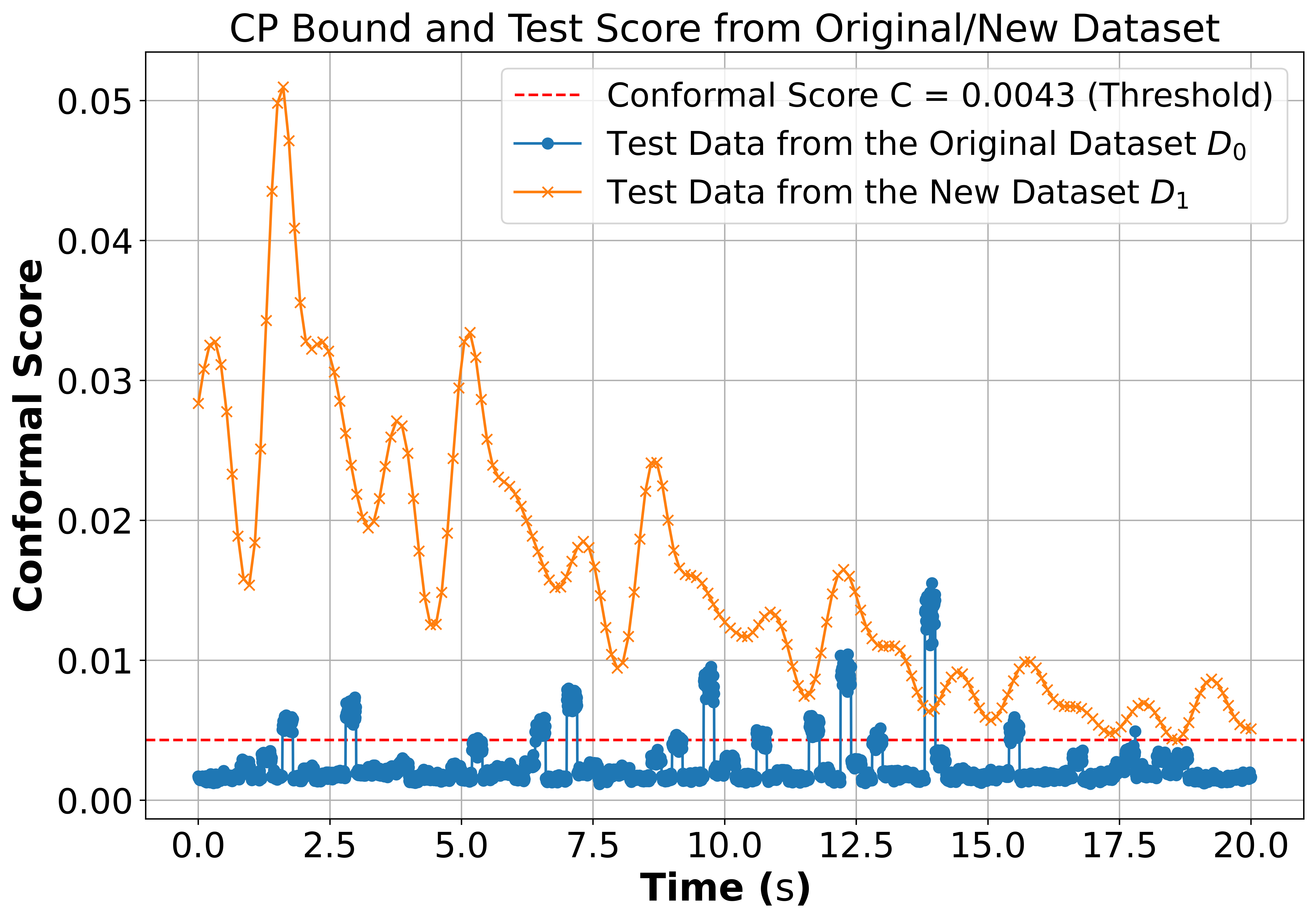}
  \vspace{-0.2cm}\caption{The nonconformity score on observation $D_{1}$ of DNN trained on $D_{0}$ is significantly higher than the conformal quantile \(C\), which indicates the existence of OOD. On the other hand, the majority of the sample scores from $D_{0}$ are lower than \(C\) to demonstrate the ID case.}
  \label{fig:CP_Detect}\vspace{-0.3cm}
\end{figure}
\begin{figure}[H]
  \centering
    \includegraphics[width=0.95\linewidth]{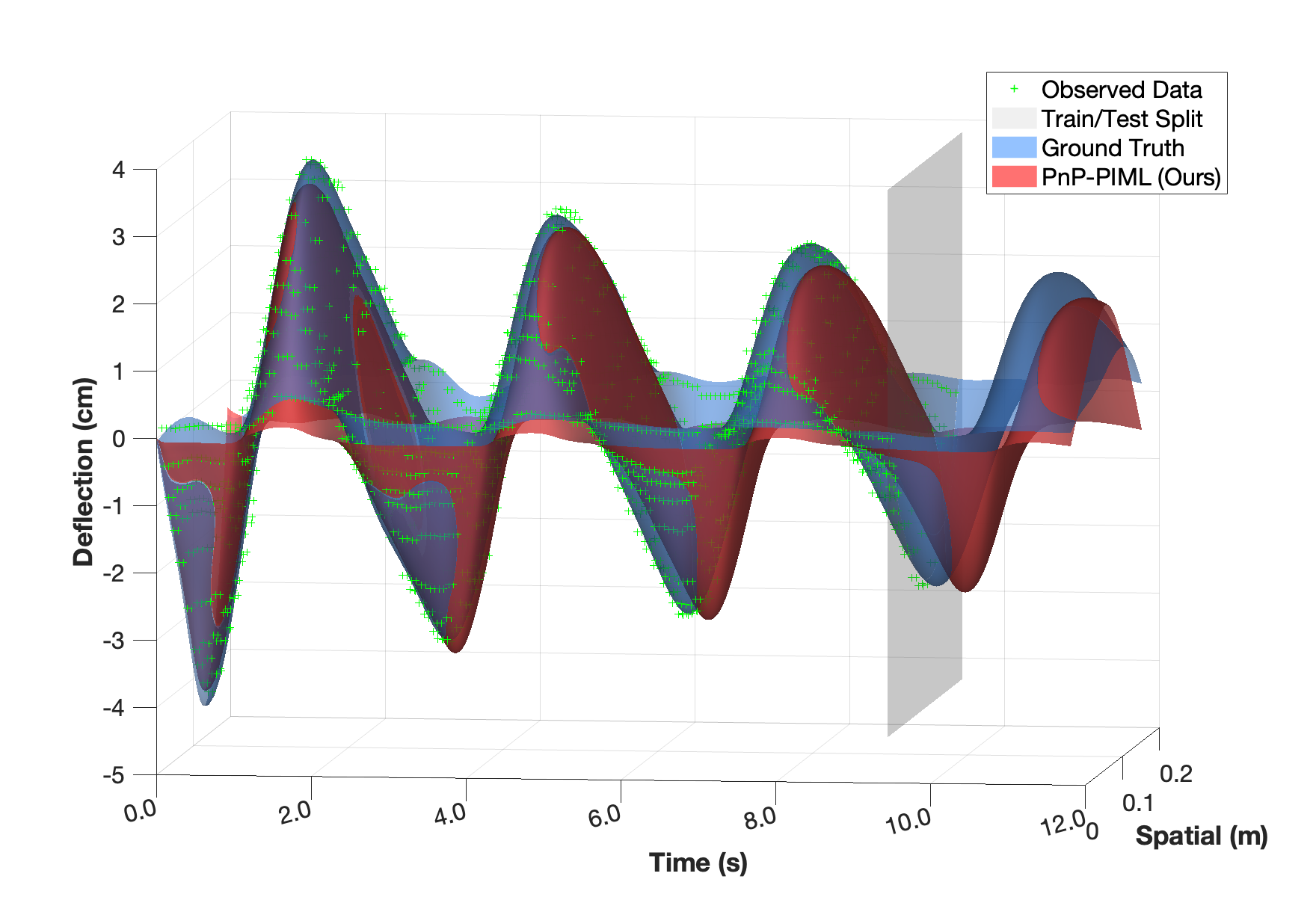}
  \caption{Proposed PnP-PIML approach, in which the physics-consistent GP-dPHS (red) generalizes well on unseen test data marked by gray plane, leading to improved accuracy.}
  \label{fig:results}\vspace{-0.2cm}
\end{figure}
\begin{figure}[H]
  \centering
  \vspace{-0.4cm}
      \includegraphics[width=0.8\linewidth]{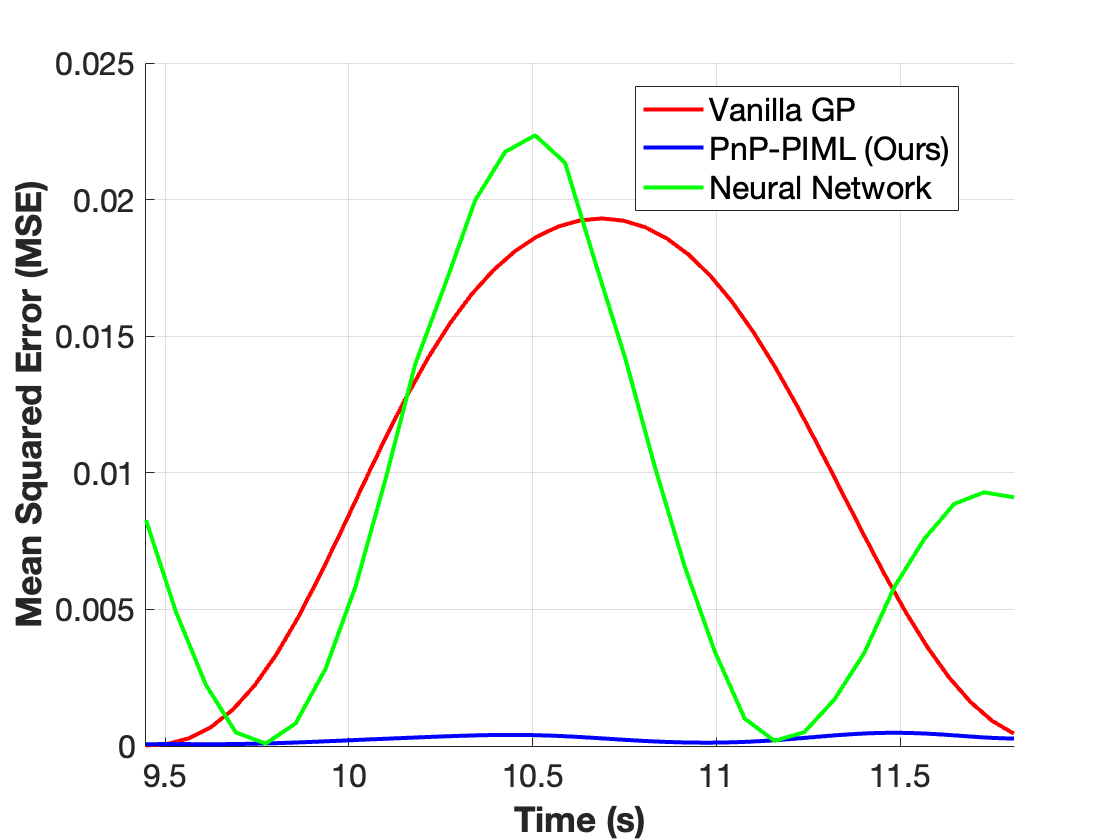}
  \vspace{-0.2cm}
  \caption{The Mean Squared Error (MSE) over time calculated on the normalized predictions, where our method (blue) performs consistently better than both baselines over the entire testing time span.}
  \label{fig:GP-dPHS_results}\vspace{-0.3cm}
\end{figure}

In the presence of OOD scenarios, the quality of the nominal predictor drops, which is noticed by the conformal prediction, see \cref{fig:CP_Detect}. Recall the score calculation in \ref{sec: CP}. A higher score $R^{new}$ represents higher prediction errors. When this score is greater than the threshold quantified by $C$, then we can say that under \( \delta = 10\% \) failure probability, this new test input is an outlier. After detecting that the nominal predictor is out-of-distribution, we switch the GP-dPHS model for prediction. As evaluation of its prediction quality, we compare it against a retrained nominal and a vanilla Gaussian Process. We train the nominal predictor, vanilla GP, and the proposed GP-dPHS model using the observed outlier data. The dataset consists of 234 time frames, each containing 50 spatial points. We use the first 80\% frames of the test points as the training set, and the remaining 20\% as the test set to evaluate model performance. For the GP-dPHS, we optimize the unknown damping factor by an Adam optimizer to $c=0.03$. The results of the prediction compared to the ground truth are presented in~\cref{fig:results,fig:GP-dPHS_results}. It can be seen, that the non-physics informed approaches (DNN and vanilla GP~\cref{fig:results}) fail to accurately predict the motion of the spring whereas our GP-dPHS captures the underlying physics and makes more accurate predictions.
  \vspace{-0.4cm}




\section{Conclusion}
In this paper, we propose the PnP-PIML framework to increase the accuracy of predictions in the case that the test data is out-of-distribution. We use conformal prediction to identify outliers and, if detected, switch from the nominal predictor to a physics-informed model. Our approach ensures the physical fidelity of the predictions, demonstrates robust generalization with limited observations, and incorporates uncertainty quantification. It can be seamlessly integrated into any existing data-driven framework to enhance performance in scenarios involving complex outlier dynamics.

\bibliographystyle{IEEEtranN}
\bibliography{mybib}

\end{document}